\documentclass{article}
\PassOptionsToPackage{numbers, compress}{natbib}
\usepackage[main,final]{neurips_2026}
\usepackage[utf8]{inputenc}
\usepackage[T1]{fontenc}
\usepackage{hyperref}
\usepackage{url}
\usepackage{booktabs}
\usepackage{amsfonts}
\usepackage{amsmath}
\usepackage{amssymb}
\usepackage{nicefrac}
\usepackage{microtype}
\usepackage{xcolor}
\usepackage{graphicx}

\title{Should Service Requirements Be Scalarized or Constrained When Minimizing Costs?\\ A Case Study on HVAC Control}

\author{%
  Tai-You Chen \quad Mihael Mohor\v{c}i\v{c} \quad Bla\v{z} Bertalani\v{c} \quad Carolina Fortuna \\
  Department of Communication Systems, Jo\v{z}ef Stefan Institute, Ljubljana, Slovenia \\
  \texttt{575583114@qq.com} \\
  \texttt{\{miha.mohorcic, blaz.bertalanic, carolina.fortuna\}@ijs.si}
}

\begin{document}
\maketitle

\begin{abstract}
Automated decision systems, from learned policies to model-based optimizers, often minimize costs such as
energy or resource use while meeting a service requirement, for example the quality of service (QoS)
specified in a service level agreement (SLA), a comfort band, or a safety margin. The requirement can be
\emph{scalarized} with the costs into a single objective, with a fixed or an adaptively tuned (Lagrangian)
weight, or it can be imposed as a \emph{constraint}. We study this choice in the Smart Buildings
Challenge, in which a single policy controls the heating, ventilation and air conditioning (HVAC) of
previously unseen buildings and is evaluated by a weighted sum of comfort, energy cost and carbon
emissions. With all other components fixed, we compare the two treatments in model-based optimization,
represented by model predictive control (MPC) with the requirement as a hard or a soft
($\ell_1$-penalized) constraint, and in learned policies, represented by proximal policy optimization
(PPO) with a fixed or a Lagrangian penalty in the reward or a projection layer on the action. In MPC, the
soft constraint reproduces the hard one once its weight exceeds the requirement's shadow price, which the
challenge's comfort weight exceeds approximately fourfold; below this weight, comfort is sacrificed, and
under a metric that values comfort below its shadow price the soft constraint outperforms the hard one by
up to 28\,\%. A weight tuned on the training buildings is 3.9\,\% worse than the hard constraint on 48
unseen buildings, because the shadow price varies across buildings. In PPO, the projected policy satisfies
the requirement from the outset and leads early in training; with longer training, the Lagrangian policy overtakes it while the projected
policy deteriorates, and the fixed penalty does not learn to satisfy the requirement. We discuss when each treatment
is preferable and when a requirement cannot be enforced by projection.
\end{abstract}

\section{Introduction}
Many automated decision systems pursue two kinds of objectives: \emph{costs} to be minimized, such as
energy, expenditure, latency or resource use, and \emph{requirements} to be met, such as a comfort band,
the latency and reliability specified in an SLA, a safety distance, or harmlessness of a language model's
responses. Examples include data-center cooling under equipment temperature limits~\cite{lazic2018data},
network resource allocation under per-class QoS targets~\cite{polese2023coloran}, robot control under
safety constraints~\cite{ji2023safety}, and language model alignment under harmlessness
constraints~\cite{dai2024safe}. These problems share a common structure: limited resources are allocated
to several demands such that the requirements are met at minimum cost.

Several methods have been proposed for such problems. The most common is \emph{weighted
scalarization}, in which the requirement violation is added to the cost as a weighted penalty; it is
widely used in reinforcement learning (RL) for building and network control~\cite{wei2017deep,wang2020reinforcement}
and in benchmark definitions~\cite{ji2023safety,goldfeder2024smart}. Multi-objective RL learns
Pareto fronts or preference-conditioned policies~\cite{roijers2013survey,abels2019dynamic,hayes2022practical},
and lexicographic methods assign priorities and tolerances to the objectives~\cite{skalse2022lexicographic}.
Constrained Markov decision processes bound an expected cumulative cost and adapt its price during
training through a Lagrange multiplier~\cite{altman1999constrained,achiam2017constrained,tessler2019reward,stooke2020responsive}.
Other methods constrain the action at execution time, through shields~\cite{alshiekh2018safe}, control
barrier functions~\cite{ames2019control}, safety and differentiable optimization
layers~\cite{dalal2018safe,amos2017optnet,donti2021dc3}, or constrained model predictive control
(MPC)~\cite{borrelli2017predictive,oldewurtel2012use}; see~\cite{garcia2015comprehensive,gu2024review} for
surveys. These methods fall into two families. With per-step cost $\kappa_t$ and requirement violation $v_t$,
\begin{align}
  &\text{scalarized, fixed weight:} && \min_\pi\ \mathbb{E}\Big[\textstyle\sum_t \big(\kappa_t + w\,v_t\big)\Big], \tag{1a}\label{eq:scal}\\
  &\text{scalarized, adapted weight (Lagrangian):} && \min_\pi\ \max_{\lambda\ge 0}\ \mathbb{E}\Big[\textstyle\sum_t \big(\kappa_t + \lambda\,(v_t-\bar v)\big)\Big], \tag{1b}\label{eq:lag}\\
  &\text{constrained:} && \min_{a_t}\ \kappa(s_t,a_t)\quad\text{s.t.}\quad \hat s_{t+1}(s_t,a_t)\in\mathcal{S}_{\text{req}}, \tag{2}\label{eq:con}
\end{align}
\setcounter{equation}{2}%
where $\bar v$ is a tolerated mean violation: Eq.~(\ref{eq:lag}) is the Lagrangian of the constraint
$\mathbb{E}\big[\frac1T\sum_{t=1}^{T} v_t\big]\le\bar v$ on the \emph{average} violation over an episode of $T$
steps, which single steps may exceed, whereas Eq.~(\ref{eq:con}) must hold at \emph{every} step. $\mathcal{S}_{\text{req}}$ is the set of states that meet the
requirement and $\hat s_{t+1}$ is a model prediction. Scalarization with a fixed weight, Eq.~(\ref{eq:scal}), is the simplest and most general
approach, but the weight must be chosen and its appropriate value may depend on the setting. An adapted
weight, Eq.~(\ref{eq:lag}), removes the manual choice but enforces the requirement only in expectation, admits
violations during learning, and may oscillate~\cite{tessler2019reward,stooke2020responsive}. A constraint,
Eq.~(\ref{eq:con}), is enforced at every step, but requires a model that maps the requirement, usually
defined on the state, to a constraint on the action.

Optimization theory relates the two families: with an $\ell_1$ penalty, the penalized and the
constrained problems have the same solutions once the weight exceeds the constraint's shadow price,
i.e.\ its Lagrange multiplier~\cite{boyd2004convex,zangwill1967nonlinear,han1979exact,nocedal2006numerical};
the same holds for soft constraints in MPC~\cite{kerrigan2000soft}. In machine learning, however,
requirements typically act on outcomes rather than actions, shadow prices vary across settings, and
learned controllers face exploration, sample efficiency and transfer. Moreover, many benchmarks use a
weighted sum as the evaluation metric, which suggests, without prescribing, a penalty-based treatment.
We therefore ask when it is worth constraining a requirement that the metric scalarizes.

We study this question in the NeurIPS 2026 Smart Buildings Challenge~\cite{goldfeder2026smart}, in which
a single policy controls the HVAC of unseen buildings and is scored by a weighted sum of comfort (0.6),
energy cost (0.2) and carbon emissions (0.2)~\cite{goldfeder2024smart}. Our contributions are:
\begin{itemize}
  \item A testbed in which only the treatment of the requirement varies, for model-based optimization
        (MPC with a soft or a hard constraint, identical model and solver) and for learned policies (PPO
        with a fixed penalty, a Lagrangian penalty or a projection layer, identical network and training length).
  \item For MPC, evidence that the shadow price decides: the soft constraint reproduces the hard one
        above it, sacrifices comfort below it, outperforms the hard constraint when the metric values
        comfort below it, and does not transfer to unseen buildings when tuned on the training ones
        (Sec.~\ref{sec:mpc-results}).
  \item For PPO, a comparison of learning under the three treatments: the projection satisfies the
        requirement from the outset, the Lagrangian overtakes it with longer training while the projected learner deteriorates, and the fixed
        penalty does not learn the requirement (Sec.~\ref{sec:rl-results}).
\end{itemize}

\section{Related works}\label{sec:related}
\textbf{Challenge and simulator.} The Smart Buildings Challenge~\cite{goldfeder2026smart} requires a single
policy that controls previously unseen commercial buildings; its control suite and
simulator~\cite{goldfeder2024smart} provide the physics, the reward and the scenario generator used in this
work. \textbf{Building control.} Building climate control is an established testbed for both kinds of
controller: MPC is the standard model-based approach~\cite{oldewurtel2012use,afram2014theory,drgona2020all},
typically based on grey-box thermal models~\cite{bacher2011identifying}; deep RL has been applied with
weighted rewards~\cite{wei2017deep,zhang2019whole,wang2020reinforcement}, and RL environments such as
CityLearn~\cite{vazquezcanteli2019citylearn} and Sinergym~\cite{jimenezraboso2021sinergym} define weighted
objectives. In energy-system competitions, optimization-based controllers have been competitive with or
superior to pure RL~\cite{nweye2024chesca,khattar2023winning}. \textbf{Requirements as penalties or
constraints.} The treatments compared here---fixed and Lagrangian penalties, projection, and hard and soft
constraints in MPC---are reviewed in the introduction; surveys of safe RL~\cite{garcia2015comprehensive,gu2024review}
and the exact-penalty literature~\cite{zangwill1967nonlinear,han1979exact,kerrigan2000soft} provide the
theoretical background.

\section{Problem}\label{sec:problem}
\textbf{Task.} A building consists of zones, each served by a variable-air-volume (VAV) box with a damper
and a hot-water reheat coil, and of air handling units (AHUs) that supply air to groups of zones; a boiler
and a pump supply the coils. Every 5\,minutes the controller observes the zone temperatures and comfort
bands, the outside temperature and the plant sensors, and sets, per AHU, the supply-air temperature, the
static pressure and the recirculation share, per zone the damper and reheat-valve positions, and the
hot-water temperature and pump pressure. Heat is delivered to a zone only through the supply air.
\textbf{Metric.} The per-step reward is $r_t = -(0.6\,c_t + 0.2\,e_t + 0.2\,g_t)$, with normalized energy cost
$e$, normalized carbon emissions $g$ and comfort shortfall $c$, the occupancy-weighted share of lost
productivity, which is linear in the deviation $\delta$ outside a zone's comfort band for $\delta\le 5$\,K
and quadratic beyond~\cite{goldfeder2024smart}. The score is the mean of $|\bar r|$ over scenarios (lower
is better). For deviations up to 5\,K, comfort therefore enters as an $\ell_1$ penalty with an implicit
weight.
\textbf{Requirement.} We regard ``every zone temperature within its comfort band'' as the requirement and
energy and carbon as the costs. Because the requirement is defined on the outcome (the next temperature)
rather than on the action, any controller that treats it as a constraint requires a model that predicts
the outcome of an action.
\textbf{Scenarios and protocol.} T30 denotes the 30 training scenarios of the challenge
dataset~\cite{goldfeder2026smart} (6 buildings $\times$ 5 climates, 30 days, 8\,640 steps each). V48
denotes 48 buildings that we generated with the challenge's generator from unused seeds, one per weather
preset, including a climate absent from T30. V48 uses seeds 9000--9067 of the generator. All rollouts use the
organizers' evaluation loop, all comparisons are paired on identical scenarios, and the simulator is
deterministic.

\section{Model-based optimization}\label{sec:mpc}
\begin{figure}[t]
  \centering
  \includegraphics[width=\linewidth]{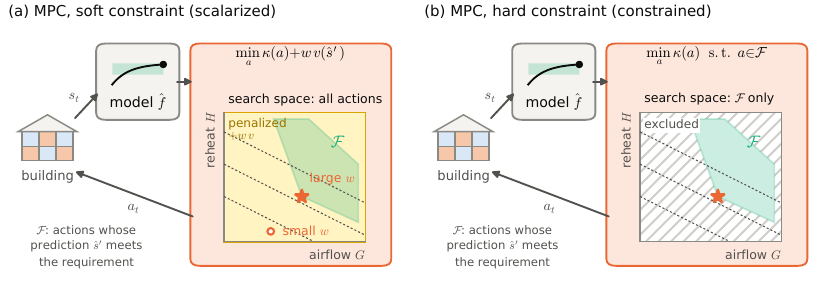}
  \caption{Model-based optimization (MPC) and the point at which the treatment of the requirement acts
  (orange). The model predicts the outcome of each action; $\mathcal{F}$ is the set of actions predicted to
  meet the requirement. (a) Soft constraint: the search covers all actions, and actions outside
  $\mathcal{F}$ are penalized; with a large weight $w$ the optimum coincides with (b), with a small weight
  it lies outside $\mathcal{F}$. (b) Hard constraint: the search is restricted to $\mathcal{F}$, and the
  cheapest action lies on its boundary.}
  \label{fig:mpc-frame}
\end{figure}
A model-based optimizer predicts the outcomes of candidate actions with a model and selects an action by
solving an optimization problem over these predictions. We use MPC, the standard representative of this
class and the established model-based approach to building control~\cite{oldewurtel2012use,afram2014theory,drgona2020all}.

\subsection{Soft and hard constraints}\label{sec:mpc-method}
At every step, MPC plans actions $a_{t},\dots,a_{t+N-1}$ over a horizon $N$ using the model predictions
$\hat s_{t+k+1} = \hat f(\hat s_{t+k}, a_{t+k})$, applies the first action, and re-plans at the next step.
With modeled cost $\kappa$ and requirement violation $v$ of a predicted state, the requirement enters the
problem as a soft or a hard constraint (Fig.~\ref{fig:mpc-frame}):
\begin{align}
  &\text{soft (scalarized):} && \min_{a_{t:t+N-1}}\ \textstyle\sum_{k=0}^{N-1} \big[\kappa(a_{t+k}) + w\, v(\hat s_{t+k+1})\big], \label{eq:mpc-soft}\\
  &\text{hard (constrained):} && \min_{a_{t:t+N-1}}\ \textstyle\sum_{k=0}^{N-1} \kappa(a_{t+k})\quad\text{s.t.}\quad \hat s_{t+k+1}\in\mathcal{S}_{\text{req}}\ \ \forall k. \label{eq:mpc-hard}
\end{align}
Eq.~(\ref{eq:mpc-soft}) is the MPC form of the fixed-weight scalarization in Eq.~(\ref{eq:scal}), and
Eq.~(\ref{eq:mpc-hard}) the MPC form of the constraint in Eq.~(\ref{eq:con}). The adapted weight of
Eq.~(\ref{eq:lag}) is not used in MPC: tuning $w$ by dual ascent would drive it to the shadow price,
which we measure directly instead (Sec.~\ref{sec:mpc-results}).
With an $\ell_1$ violation, Eq.~(\ref{eq:mpc-soft}) has the solution of Eq.~(\ref{eq:mpc-hard}) once $w$ exceeds
the constraint's \emph{shadow price}, i.e.\ its Lagrange multiplier in Eq.~(\ref{eq:mpc-hard}), the marginal cost of
meeting the requirement expressed in the units of $w$~\cite{kerrigan2000soft}; below it, the soft problem relaxes the requirement
wherever meeting it costs more than $w$. If Eq.~(\ref{eq:mpc-hard}) is infeasible, the hard constraint must be
relaxed; we then minimize the violation first and the cost second, which corresponds to
Eq.~(\ref{eq:mpc-soft}) with $w\to\infty$.

\subsection{Experimental setup}\label{sec:mpc-setup}
\textbf{Model.} Each zone is described by a grey-box first-order heat balance~\cite{bacher2011identifying},
discretized implicitly over one 5-minute step,
\begin{equation}
  T' = \frac{T + d + \gamma\,(G\,T_s + H)}{1 + \gamma\,G},\label{eq:zone}
\end{equation}
where $T$ is the zone temperature, $G$ the airflow expressed as a thermal conductance (W\,K$^{-1}$),
$T_s$ the supply-air temperature, $H$ the coil heat, $\gamma$ the zone's response and $d$ its free drift
(envelope, solar and internal gains); $\gamma$ and $d$ are identified online by recursive least squares.
Eq.~(\ref{eq:zone}) predicts the next temperature for a given action; conversely, for fixed $T_s$
the band condition $\ell\le T'\le u$ is linear in $(G,H)$, so the actions that reach $\ell$ or $u$
exactly are available in closed form.
\textbf{Horizon and problem.} We use a horizon of one step, $N=1$; multi-step effects (night and weekend
set-back, morning warm-up, hot-water plant) follow the same fixed rules in all variants. Over the AHU set
points and the zone actions, the one-step forms of Eqs.~(\ref{eq:mpc-soft}) and~(\ref{eq:mpc-hard}) are
\begin{align}
  &\text{MPC-S:} && \min\ \kappa + w \textstyle\sum_i \big[(\ell_i - T'_i)_+ + (T'_i-u_i)_+\big],\label{eq:mpcs}\\
  &\text{MPC-H:} && \min\ \kappa \quad \text{s.t.}\quad \ell_i \le T'_i \le u_i\ \ \forall i,\label{eq:mpch}
\end{align}
where $\kappa$ is the modeled energy and carbon cost of the step (fan, chiller, coil heat) and $w$ is
expressed in W\,K$^{-1}$ per zone.
\textbf{Solver.} The product $G\,T_s$ renders the problem bilinear, and unit on/off and economizer
decisions make it mixed-integer. We enumerate the AHU-level choices (supply-air temperature on a 0.5\,K
grid, economizer mode, off); for each choice the problem separates into one two-variable problem per
zone, which is solved exactly in $H$ (the optimum lies in $\{0,H_{\max},H(T'{=}\ell),H(T'{=}u)\}$) and
over a grid in $G$ augmented with the break points of the piecewise-linear objective (121 levels; 11
levels select identical actions, i.e.\ the optimum lies on the break points). MPC-H and MPC-S share the
solver, enumeration, model and comfort band and differ only in Eq.~(\ref{eq:mpcs}) versus Eq.~(\ref{eq:mpch}).
\textbf{Implementation.} The zone model, its identification and the AHU search reuse open-source code
published by challenge participants on Kaggle (attribution in our code). A 30-day rollout takes 5--20
minutes on one CPU core.

\subsection{Results}\label{sec:mpc-results}
\textbf{Coincidence of the formulations.} Above a threshold weight, the two formulations coincide
(Table~\ref{tab:mpc}). On T30, MPC-S with $w=10^5$ and $w=10^4$ is within 0.2\,\% of MPC-H, with the same
comfort loss, as the exact-penalty property predicts; the agreement of MPC-H and MPC-S with $w=10^5$ is
therefore expected rather than coincidental. At $w=3000$, MPC-S trades comfort (six times the comfort
loss of MPC-H) for 3\,\% lower energy and carbon, and its score is 7.5\,\% worse; at $w=1000$ it
effectively abandons the requirement (+1\,573\,\%). On the 48 unseen buildings the threshold increases:
$w=10^5$ still matches MPC-H, whereas $w=10^4$ is 3.9\,\% worse and $w=3000$ sacrifices comfort in almost
every building (+379\,\%).
\begin{table}[t]
  \caption{MPC-H versus MPC-S (identical model and solver), paired comparison. Score, comfort
  ($0.6\,c_t$) and energy plus carbon ($0.2\,e_t+0.2\,g_t$) per step ($10^{-3}$, lower is better); the
  last two columns sum to the score.}
  \label{tab:mpc}
  \centering\small
  \begin{tabular}{lrrrrrr}
    \toprule
    & \multicolumn{3}{c}{T30 (30 scenarios)} & \multicolumn{3}{c}{V48 (48 unseen buildings)} \\
    \cmidrule(lr){2-4}\cmidrule(lr){5-7}
    & score & comfort & en.+carb. & score & comfort & en.+carb. \\
    \midrule
    MPC-H & 8.50 & 0.16 & 8.34 & 11.06 & 0.59 & 10.47 \\
    MPC-S, $w=10^5$ & 8.50 & 0.16 & 8.34 & 11.06 & 0.59 & 10.47 \\
    MPC-S, $w=10^4$ & 8.48 & 0.16 & 8.32 & 11.49 & 1.10 & 10.39 \\
    MPC-S, $w=3000$ & 9.14 & 1.04 & 8.10 & 52.97 & 44.75 & 8.21 \\
    MPC-S, $w=1000$ & 142.2 & 137.5 & 4.64 & 630.9 & 629.0 & 1.89 \\
    \bottomrule
  \end{tabular}
\end{table}
\textbf{Role of the shadow price.} We estimate the shadow price in the units of the metric by a finite
difference, i.e.\ as an average over an interval of weights. Between $w=10^4$ and $w=3000$ on T30
(Table~\ref{tab:mpc}), the weighted comfort loss increases by $1.04-0.16=0.88\cdot10^{-3}$, i.e.\ the comfort
loss $c$ by $\Delta c=0.88/0.6\approx1.47\cdot10^{-3}$, whereas the weighted energy and carbon decrease by only
$\Delta\kappa=8.32-8.10=0.22\cdot10^{-3}$. One unit of comfort loss thus saves $\Delta\kappa/\Delta c\approx0.15$
units of weighted cost; this is the shadow price in the units of the metric, and the metric's comfort weight
of 0.6 exceeds it by a factor of 3.9 (computed from unrounded values).
The metric therefore values the requirement well above the cost of meeting it, which is precisely the
regime in which a constraint performs as well as a well-tuned penalty without requiring a weight.
\begin{figure}[t]
  \centering
  \includegraphics[width=0.55\linewidth]{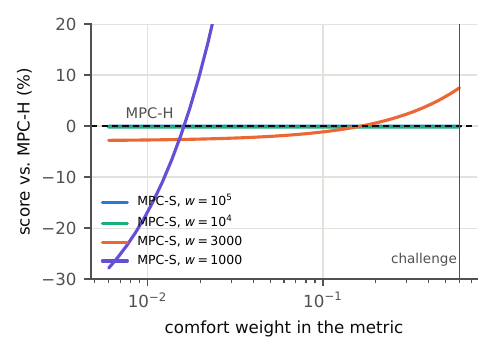}
  \caption{The T30 runs re-scored with a reduced comfort weight in the metric (energy and carbon retain
  0.2 each). Score of MPC-S relative to MPC-H (negative values: the penalty is better). The challenge
  weight is 0.6 (vertical line); below approximately 0.15, a soft constraint with a suitable $w$ is better. The crossover equals the shadow
  price estimated from Table~\ref{tab:mpc} (Sec.~\ref{sec:mpc-results}).}
  \label{fig:reweight}
\end{figure}
\textbf{When the penalty is preferable.} If the metric values the requirement below its shadow price,
meeting it exactly is no longer optimal. We re-score the same runs with reduced comfort weights; this is
valid because comfort, energy and carbon are recorded separately and the controllers do not observe the
metric (Fig.~\ref{fig:reweight}). With the challenge weight (0.6) and with 0.2, MPC-H is as good as the
best MPC-S; with 0.06, MPC-S with $w=3000$ is 1.8\,\% better, with 0.02 it is 2.5\,\% better, and with
0.006, MPC-S with $w=1000$ is 28\,\% better. The crossover is the shadow price: with comfort weight $\alpha$,
the soft constraint is better when $\alpha\,\Delta c<\Delta\kappa$, i.e.\ for $\alpha<0.15\approx0.6/3.9$. A constraint is thus the better treatment when the requirement is worth more than the cost of
meeting it, and a penalty is better otherwise, provided that its weight matches the metric.
\textbf{Transfer.} The weight selected on the training buildings does not transfer. On T30, the best soft
constraint is $w=10^4$ ($-0.2$\,\% relative to MPC-H); on V48, the same weight is 3.9\,\% worse, exceeds
MPC-H by more than 1\,\% in 6 of 48 buildings, and differs across climates (Phoenix $+11.6$\,\%,
Minneapolis $+4.9$\,\%, Fairbanks $+3.8$\,\%; Atlanta, Denver and Singapore unchanged). Since the shadow
price varies with building and climate, a soft constraint is safe on unseen buildings only with a weight
well above every shadow price, in which case it coincides with the hard constraint, which requires no weight.

\textbf{Hidden test set.} On the challenge's hidden test buildings (Kaggle public leaderboard; lower is
better), MPC-H scores 0.01185 and MPC-S with $w=10^4$ scores 0.01184, i.e.\ practically identical, as on T30. The leaderboard reports a
single aggregate per submission, so this comparison is unpaired and serves as supplementary evidence.

\section{Learned policies}\label{sec:rl}
A learned policy maps the observation directly to an action, without solving an optimization problem at
run time. We use proximal policy optimization (PPO)~\cite{schulman2017proximal}, a widely used
policy-gradient method, as the representative learner.

\begin{figure}[t]
  \centering
  \includegraphics[width=\linewidth]{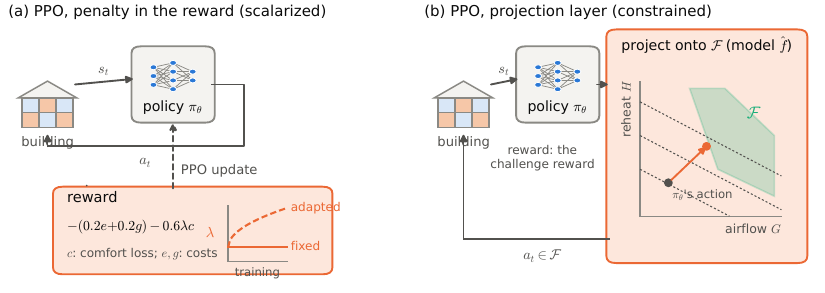}
  \caption{Learned policy (PPO) and the point at which the treatment of the requirement acts (orange).
  (a) Penalty in the reward: the comfort loss $c$ is weighted against the costs with a fixed weight
  $\lambda$ or with a weight adapted during training (Lagrangian). (b) Projection layer: the policy's
  action is replaced by the nearest action in $\mathcal{F}$, the set predicted by a model to meet the
  requirement; the reward is the challenge reward.}
  \label{fig:ppo-frame}
\end{figure}
\subsection{Penalties and projection}\label{sec:rl-method}
The requirement can enter either the learning signal or the action (Fig.~\ref{fig:ppo-frame}). Two
treatments scalarize it in the reward and one constrains the action. \emph{Fixed penalty},
Eq.~(\ref{eq:scal}): the reward is the challenge reward, so the requirement carries a fixed weight.
\emph{Lagrangian}, Eq.~(\ref{eq:lag})~\cite{tessler2019reward}: the reward is $-(0.2\,e_t+0.2\,g_t)-\lambda\cdot0.6\,c_t$, which equals
the challenge reward at $\lambda=1$, and $\lambda$ is increased while the training-time violation $\bar c$,
the mean of $0.6\,c_t$ over the iteration, exceeds a threshold $\bar c_{\max}$,
$\lambda\leftarrow\mathrm{clip}(\lambda+\eta(\bar c-\bar c_{\max}),0,\lambda_{\max})$. In terms of
Eq.~(\ref{eq:lag}), $v_t=0.6\,c_t$, $\bar v=\bar c_{\max}$, and $\bar c$ is the sample estimate of the average
violation: the multiplier constrains the average, not each step.
\emph{Projection}, Eq.~(\ref{eq:con})~\cite{dalal2018safe}: before it is applied, the policy's action is
projected onto the set of actions that a model predicts to meet the requirement; the projection is part
of the environment, and the reward is the challenge reward. Only the projection uses a model to
\emph{enforce} the requirement.

\subsection{Experimental setup}\label{sec:rl-setup}
The policy determines the zone-level airflow and reheat; the AHU set points are selected by the same
search as in the MPC experiments, so that the learner faces exactly the per-zone allocation problem.
\textbf{Policy.} An actor-critic network shared by all zones (two hidden layers of 32 tanh units) maps nine
zone features to a Gaussian distribution over pre-sigmoid airflow and reheat shares $(u,v)\in[0,1]^2$
(initial log standard deviation $-0.5$). The features (predicted shortfall and excess from
Eq.~(\ref{eq:zone}), distances to the band, $T_s$, outside temperature, comfort mode, drift and a bias) are
derived from the identified zone model, so all three learners receive the same model information. Sharing
the network makes the policy applicable to buildings of any size and yields one transition per zone and
step; all zones of a step receive that step's reward.
\textbf{Treatment parameters.} Lagrangian: $\lambda_0=1$ (the challenge weights), $\eta=20$,
$\bar c_{\max}=3\cdot10^{-4}$ per step, $\lambda_{\max}=100$. Projection: Euclidean projection, in the normalized
coordinates $(u,v)$, onto the polygon of actions that Eq.~(\ref{eq:zone}) predicts to keep the zone within its
band (seven half-planes: the two band limits and the bounds of $u$ and $v$ and of the coil heat), computed with
40 iterations of Dykstra's algorithm; if the polygon is empty, the policy's action is applied unchanged.
\textbf{Training and evaluation.} All three learners use CleanRL defaults~\cite{huang2022cleanrl} (clip
0.2, GAE $\lambda=0.95$, $\gamma=0.97$, 4 epochs, minibatches of 8\,192 zone transitions, Adam, learning
rate $10^{-3}$); 16 parallel workers each simulate one day per iteration on a scenario drawn from six T30
scenarios. Each treatment is trained for $4\cdot10^5$ environment steps ($\approx$46 building-months) with
three seeds, and in two further runs for $1.2\cdot10^6$ steps, which we then continued to $3\cdot10^6$ steps
from the final checkpoint with a re-initialized optimizer (the Lagrange multiplier is carried over). Every $\approx$10 iterations, the mean
action is evaluated on four held-out T30 building--climate pairs (Atlanta, Phoenix, Fairbanks, Denver)
over their first 7 days. A run of $4\cdot10^5$ steps takes approximately 1.5 hours on 16 CPU cores.

\subsection{Results}\label{sec:rl-results}
\begin{figure}[t]
  \centering
  \includegraphics[width=\linewidth]{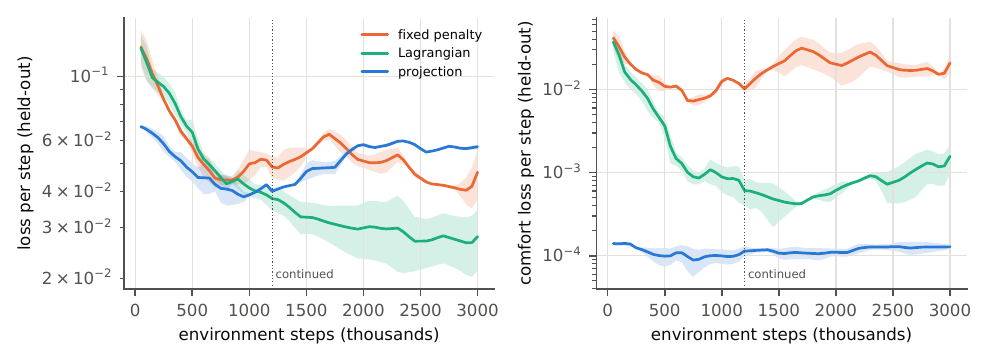}
  \caption{PPO with the three treatments (two scalarized, one constrained), evaluated during training on
  four held-out scenarios (mean action, first 7 days). Left: loss per step; right: comfort loss per step.
  Lines: mean of two runs per treatment, trained to $1.2\cdot10^6$ steps and continued from the checkpoint
  (dotted line) to $3\cdot10^6$ steps; bands: range over the two runs.}
  \label{fig:ppo}
\end{figure}
\begin{table}[t]
  \caption{PPO on four held-out scenarios (first 7 days; loss and comfort loss per step, $10^{-3}$) after
  $\approx5\cdot10^4$, $1.5\cdot10^5$ and $4\cdot10^5$ environment steps (mean over three seeds) and after
  $1.2\cdot10^6$ and $3\cdot10^6$ steps (mean of two runs; continued from $1.2\cdot10^6$ steps).}
  \label{tab:ppo}
  \centering\small\setlength{\tabcolsep}{3.2pt}
  \begin{tabular}{lrrrrrrrrrr}
    \toprule
    & \multicolumn{2}{c}{$5\cdot10^4$} & \multicolumn{2}{c}{$1.5\cdot10^5$} & \multicolumn{2}{c}{$4\cdot10^5$} & \multicolumn{2}{c}{$1.2\cdot10^6$} & \multicolumn{2}{c}{$3\cdot10^6$} \\
    \cmidrule(lr){2-3}\cmidrule(lr){4-5}\cmidrule(lr){6-7}\cmidrule(lr){8-9}\cmidrule(lr){10-11}
    Treatment & loss & comf. & loss & comf. & loss & comf. & loss & comf. & loss & comf. \\
    \midrule
    Fixed penalty & 92.3 & 21.9 & 68.8 & 11.3 & 49.9 & 5.7 & 48.7 & 10.1 & 47.1 & 21.2 \\
    Lagrangian & 93.2 & 21.3 & 67.1 & 8.7 & 46.6 & 2.2 & 37.8 & 0.6 & 28.0 & 1.6 \\
    Projection & 61.4 & 0.14 & 54.0 & 0.12 & 43.7 & 0.10 & 40.0 & 0.12 & 57.2 & 0.13 \\
    \bottomrule
  \end{tabular}
\end{table}
\emph{Early training.} At the first evaluation ($\approx5\cdot10^4$ steps), the projected learner keeps
the comfort loss at $1.4\cdot10^{-4}$ per step, whereas the fixed-penalty and Lagrangian learners incur
approximately 150 times more, and their total loss is approximately 1.5 times higher (0.092 and 0.093
versus 0.061).
\emph{Intermediate training.} Up to $1.5\cdot10^5$ steps, approximately half of the improvement of the
scalarized learners stems from comfort (fixed penalty: 0.011 of 0.024; Lagrangian: 0.013 of 0.026); these
learners first learn to meet the requirement, whereas the projected learner, which cannot improve comfort,
improves energy and carbon during the same period.
\emph{After $4\cdot10^5$ steps,} the ranking is projection (0.044), Lagrangian (0.047) and fixed penalty
(0.050), but the seeds overlap (projection 0.033--0.052); the treatments are clearly separated only by the
comfort loss ($1.0\cdot10^{-4}$, $2.2\cdot10^{-3}$ and $5.7\cdot10^{-3}$).
\emph{Longer training} changes the ranking (two runs per treatment; Fig.~\ref{fig:ppo}). Up to
$1.2\cdot10^6$ steps, the Lagrangian learner catches up with the projected learner (0.034 and 0.041 versus
0.040 and 0.040), and its comfort loss decreases to 4--8$\cdot10^{-4}$. In the continuation to
$3\cdot10^6$ steps, it improves further, to 0.021 and 0.035, the lowest losses of all learners. Its held-out
comfort loss, however, rises again to 1--2$\cdot10^{-3}$, although its training-time comfort loss continues
to fall. Its multiplier increased throughout training (to 61 and 50), because the training-time comfort loss
fell below the threshold only in iterations that simulated an almost unoccupied building. The projected
learner, in contrast, deteriorates from approximately $1.3\cdot10^6$ steps, to 0.058 and 0.056 at
$3\cdot10^6$ steps. The deterioration occurs in both runs and only in energy and carbon; the comfort loss
remains at 1.2--1.4$\cdot10^{-4}$. It also appears in the training-time energy and carbon (an increase of
approximately 50--80\,\%), so it is not specific to the held-out scenarios; it sets in gradually rather
than at the restart of the optimizer, and the policy's standard deviation decreases throughout, so it is
not caused by exploration. We have not identified its cause; one candidate is that the policy gradient
credits the policy's own action, whereas the environment receives the projected action, so that the
learning signal does not account for the projection. The fixed-penalty learner does not learn to meet the
requirement: its comfort loss remains at 0.7--4.4$\cdot10^{-2}$, and its loss fluctuates between 0.034 and
0.066. This is not because violations are cheap:
the challenge's comfort weight is well above the requirement's shadow price (Sec.~\ref{sec:mpc-results}),
so the penalized and constrained problems share their optimum. We conjecture that the learning signal is
the limiting factor: a single reward per step is shared by all zones, which dilutes the contribution of
each zone's violation, whereas the Lagrange multiplier (increasing from 1 to 19--30 within $4\cdot10^5$ steps) amplifies it.
\emph{Requirements of each treatment.} The projection requires a model that maps the outcome requirement
to a set of actions; it benefits from this knowledge early, but in our runs it did not improve with longer training. The fixed penalty requires only
the metric and must learn the requirement from violations. The Lagrangian requires a threshold and a step
size, and its multiplier increases while the requirement is violated, which changes the reward optimized
during training.

\textbf{Hidden test set.} On the Kaggle public leaderboard (lower is better), the PPO policies (final
checkpoint of the first $1.2\cdot10^6$-step run of each treatment) score 0.0585 with the fixed
penalty, 0.0354 with the Lagrangian and 0.0499 with the projection, in the same order as these checkpoints on the
held-out scenarios (0.053, 0.034 and 0.040); as for MPC, these are single
unpaired aggregates.

\section{Discussion and limitations}\label{sec:discussion}
\textbf{When is scalarization preferable?} When the metric weights the requirement below its shadow price,
meeting it exactly is not optimal, and a penalty with the metric's weight is the appropriate objective
(Sec.~\ref{sec:mpc-results}). A penalty is also the only option when no model can translate the
requirement into a constraint on the action, and it degrades gracefully when the requirement cannot be met
(e.g., during warm-up), where a hard constraint is infeasible and needs a fallback rule (here, minimizing the
violation first, i.e.\ $w\to\infty$; Sec.~\ref{sec:mpc-method}).
\textbf{Can every requirement be enforced by projection?} Only if a model maps actions to the outcome on
which the requirement is defined and the resulting action set is tractable. Here, the zone model yields a
polygon for each supply-air temperature; for requirements on long-term or stochastic outcomes, or with
nonconvex action sets, the projection becomes approximate or computationally expensive, and its quality is
bounded by that of the model.
\textbf{Limitations.} Our MPC uses a one-step horizon; multi-step effects (thermal mass, set-back depth)
follow fixed rules. The model is identified online and approximate, so the constraint applies to the
\emph{predicted} temperature. All results are obtained with a single simulator. The learned-policy study
uses three seeds up to $4\cdot10^5$ steps, two runs beyond, and four held-out scenarios of 7 days; the runs
beyond $1.2\cdot10^6$ steps restart the optimizer, the cause of the projected learner's deterioration is open,
and the Lagrangian results depend on the threshold and the step size.

\section{Conclusion}
Whether a service requirement should be scalarized or constrained depends on two quantities: the weight
that the metric assigns to it and its shadow price. In the Smart Buildings Challenge the weight exceeds the
shadow price approximately four times (3.9); in this regime, a model-based optimizer with a hard constraint performs as well
as the best soft constraint, requires no weight, and transfers to unseen buildings, where the shadow price
is higher and a weight tuned on the training buildings is insufficient. Under a metric that values the
requirement below its shadow price, the soft constraint is preferable. For learned policies, the treatment
shapes learning: a projection meets the requirement from the outset but requires a model, and in our runs it deteriorated
with longer training; a Lagrangian weight learns the requirement more slowly but eventually attains the
lowest loss; and a fixed penalty, despite a
weight well above the shadow price, did not learn the requirement within the training lengths we used.

\begin{ack}
The 6G-OPTICON project (6G Optimization and Constrained Intelligence for End-to-End Network Operations
across Heterogeneous Segments) has received funding from the European Union's HORIZON-JU-SNS-2025 under
Grant Agreement No.~101292327 (Research and Innovation Action). The EnerTEF project (Common European-scale
Energy Artificial Intelligence Federated Testing and Experimentation Facility) has received funding from
the European Union's HORIZON-CL5-2024 under Grant Agreement No.~101172887 (Research and Innovation Action).
\end{ack}

{\small
\bibliographystyle{plainnat}
\bibliography{references}

@InProceedings{abels2019dynamic,
  title = 	 {{Dynamic Weights in Multi-Objective Deep Reinforcement Learning}},
  author =       {Abels, Axel and Roijers, Diederik and Lenaerts, Tom and Now{\'e}, Ann and Steckelmacher, Denis},
  booktitle = 	 {Proceedings of the 36th International Conference on Machine Learning},
  pages = 	 {11--20},
  year = 	 {2019},
  editor = 	 {Chaudhuri, Kamalika and Salakhutdinov, Ruslan},
  volume = 	 {97},
  series = 	 {Proceedings of Machine Learning Research},
  publisher =    {PMLR}
}

@InProceedings{achiam2017constrained,
  title = 	 {{Constrained Policy Optimization}},
  author =       {Joshua Achiam and David Held and Aviv Tamar and Pieter Abbeel},
  booktitle = 	 {Proceedings of the 34th International Conference on Machine Learning},
  pages = 	 {22--31},
  year = 	 {2017},
  editor = 	 {Precup, Doina and Teh, Yee Whye},
  volume = 	 {70},
  series = 	 {Proceedings of Machine Learning Research},
  publisher =    {PMLR}
}

@article{afram2014theory, title={{Theory and applications of HVAC control systems – A review of model predictive control (MPC)}}, volume={72}, ISSN={0360-1323}, DOI={10.1016/j.buildenv.2013.11.016}, journal={Building and Environment}, publisher={Elsevier BV}, author={Afram, Abdul and Janabi-Sharifi, Farrokh}, year={2014}, pages={343–355} }

@article{alshiekh2018safe, title={{Safe Reinforcement Learning via Shielding}}, volume={32}, ISSN={2159-5399}, DOI={10.1609/aaai.v32i1.11797}, number={1}, journal={Proceedings of the AAAI Conference on Artificial Intelligence}, publisher={Association for the Advancement of Artificial Intelligence (AAAI)}, author={Alshiekh, Mohammed and Bloem, Roderick and Ehlers, Rüdiger and Könighofer, Bettina and Niekum, Scott and Topcu, Ufuk}, year={2018} }

@book{altman1999constrained, title={{Constrained Markov Decision Processes: Stochastic Modeling}}, ISBN={9781315140223}, DOI={10.1201/9781315140223}, publisher={Routledge}, author={Altman, Eitan}, year={2021} }

@inproceedings{ames2019control, title={{Control Barrier Functions: Theory and Applications}}, DOI={10.23919/ecc.2019.8796030}, booktitle={2019 18th European Control Conference (ECC)}, publisher={IEEE}, author={Ames, Aaron D. and Coogan, Samuel and Egerstedt, Magnus and Notomista, Gennaro and Sreenath, Koushil and Tabuada, Paulo}, year={2019}, pages={3420–3431} }

@InProceedings{amos2017optnet,
  title = 	 {{O}pt{N}et: Differentiable Optimization as a Layer in Neural Networks},
  author =       {Brandon Amos and J. Zico Kolter},
  booktitle = 	 {Proceedings of the 34th International Conference on Machine Learning},
  pages = 	 {136--145},
  year = 	 {2017},
  editor = 	 {Precup, Doina and Teh, Yee Whye},
  volume = 	 {70},
  series = 	 {Proceedings of Machine Learning Research},
  publisher =    {PMLR}
}

@article{bacher2011identifying, title={{Identifying suitable models for the heat dynamics of buildings}}, volume={43}, ISSN={0378-7788}, DOI={10.1016/j.enbuild.2011.02.005}, number={7}, journal={Energy and Buildings}, publisher={Elsevier BV}, author={Bacher, Peder and Madsen, Henrik}, year={2011}, pages={1511–1522} }

@book{boyd2004convex, title={{Convex Optimization}}, ISBN={9780511804441}, DOI={10.1017/cbo9780511804441}, publisher={Cambridge University Press}, author={Boyd, Stephen and Vandenberghe, Lieven}, year={2004} }

@inproceedings{dai2024safe,
title={{Safe {RLHF}: Safe Reinforcement Learning from Human Feedback}},
author={Josef Dai and Xuehai Pan and Ruiyang Sun and Jiaming Ji and Xinbo Xu and Mickel Liu and Yizhou Wang and Yaodong Yang},
booktitle={The Twelfth International Conference on Learning Representations},
year={2024}
}

@misc{dalal2018safe,
  doi = {10.48550/ARXIV.1801.08757},
  author = {Dalal, Gal and Dvijotham, Krishnamurthy and Vecerik, Matej and Hester, Todd and Paduraru, Cosmin and Tassa, Yuval},
  title = {{Safe Exploration in Continuous Action Spaces}},
  publisher = {arXiv},
  year = {2018},
  copyright = {arXiv.org perpetual, non-exclusive license},
  note={arXiv:1801.08757}
}

@inproceedings{donti2021dc3,
title={{DC}3: A learning method for optimization with hard constraints},
author={Priya L. Donti and David Rolnick and J Zico Kolter},
booktitle={International Conference on Learning Representations},
year={2021}
}

@article{drgona2020all, title={{All you need to know about model predictive control for buildings}}, volume={50}, ISSN={1367-5788}, DOI={10.1016/j.arcontrol.2020.09.001}, journal={Annual Reviews in Control}, publisher={Elsevier BV}, author={Drgoňa, Ján and Arroyo, Javier and Cupeiro Figueroa, Iago and Blum, David and Arendt, Krzysztof and Kim, Donghun and Ollé, Enric Perarnau and Oravec, Juraj and Wetter, Michael and Vrabie, Draguna L. and Helsen, Lieve}, year={2020}, pages={190–232} }

@article{garcia2015comprehensive,
  author  = {Javier García and Fernando Fernández},
  title   = {{A Comprehensive Survey on Safe Reinforcement Learning}},
  journal = {Journal of Machine Learning Research},
  year    = {2015},
  volume  = {16},
  number  = {42},
  pages   = {1437--1480}
}

@misc{goldfeder2024smart,
  doi = {10.48550/ARXIV.2410.03756},
  author = {Goldfeder, Judah and Dean, Victoria and Jiang, Zixin and Wang, Xuezheng and dong, Bing and Lipson, Hod and Sipple, John},
  title = {{The Smart Buildings Control Suite: A Diverse Open Source Benchmark to Evaluate and Scale HVAC Control Policies for Sustainability}},
  publisher = {arXiv},
  year = {2024},
  copyright = {Creative Commons Attribution 4.0 International},
  note={arXiv:2410.03756}
}

@misc{goldfeder2026smart,
  doi = {10.48550/ARXIV.2610.18511},
  author = {Goldfeder, Judah and Rai, Ayush and Wyder, Philippe M. and Li, Na and Dean, Victoria and Vakalis, Donna and Lin, Jiong and Dong, Bing and Rossetti, Michael and Kutz, Nathan and Arisaka, Sohei and Sipple, John and Lipson, Hod and Bengio, Yoshua},
  title = {{The Smart Buildings Challenge: Learning Foundation Models for Real-World Optimization at Scale}},
  publisher = {arXiv},
  year = {2026},
  copyright = {Creative Commons Attribution 4.0 International},
  note={arXiv:2610.18511}
}

@article{gu2024review, title={{A Review of Safe Reinforcement Learning: Methods, Theories, and Applications}}, volume={46}, ISSN={1939-3539}, DOI={10.1109/tpami.2024.3457538}, number={12}, journal={IEEE Transactions on Pattern Analysis and Machine Intelligence}, publisher={Institute of Electrical and Electronics Engineers (IEEE)}, author={Gu, Shangding and Yang, Long and Du, Yali and Chen, Guang and Walter, Florian and Wang, Jun and Knoll, Alois}, year={2024}, pages={11216–11235} }

@article{han1979exact, title={{Exact penalty functions in nonlinear programming}}, volume={17}, ISSN={1436-4646}, DOI={10.1007/bf01588250}, number={1}, journal={Mathematical Programming}, publisher={Springer Science and Business Media LLC}, author={Han, S. -P. and Mangasarian, O. L.}, year={1979}, pages={251–269} }

@article{hayes2022practical, title={{A practical guide to multi-objective reinforcement learning and planning}}, volume={36}, ISSN={1573-7454}, DOI={10.1007/s10458-022-09552-y}, number={1}, journal={Autonomous Agents and Multi-Agent Systems}, publisher={Springer Science and Business Media LLC}, author={Hayes, Conor F. and Rădulescu, Roxana and Bargiacchi, Eugenio and Källström, Johan and Macfarlane, Matthew and Reymond, Mathieu and Verstraeten, Timothy and Zintgraf, Luisa M. and Dazeley, Richard and Heintz, Fredrik and Howley, Enda and Irissappane, Athirai A. and Mannion, Patrick and Nowé, Ann and Ramos, Gabriel and Restelli, Marcello and Vamplew, Peter and Roijers, Diederik M.}, year={2022} }

@article{huang2022cleanrl,
  author  = {Shengyi Huang and Rousslan Fernand Julien Dossa and Chang Ye and Jeff Braga and Dipam Chakraborty and Kinal Mehta and João G.M. Araújo},
  title   = {{CleanRL: High-quality Single-file Implementations of Deep Reinforcement Learning Algorithms}},
  journal = {Journal of Machine Learning Research},
  year    = {2022},
  volume  = {23},
  number  = {274},
  pages   = {1--18}
}

@inproceedings{jimenezraboso2021sinergym, series={BuildSys ’21}, title={{Sinergym: a building simulation and control framework for training reinforcement learning agents}}, DOI={10.1145/3486611.3488729}, booktitle={Proceedings of the 8th ACM International Conference on Systems for Energy-Efficient Buildings, Cities, and Transportation}, publisher={ACM}, author={Jiménez-Raboso, Javier and Campoy-Nieves, Alejandro and Manjavacas-Lucas, Antonio and Gómez-Romero, Juan and Molina-Solana, Miguel}, year={2021}, pages={319–323}, collection={BuildSys ’21} }

@Inproceedings{kerrigan2000soft,
 author = {E. Kerrigan and J. Maciejowski},
 title = {{Soft constraints and exact penalty functions in model predictive control}},
 year = {2000}
}

@article{khattar2023winning, title={{Winning the CityLearn Challenge: Adaptive Optimization with Evolutionary Search under Trajectory-Based Guidance}}, volume={37}, ISSN={2159-5399}, DOI={10.1609/aaai.v37i12.26671}, number={12}, journal={Proceedings of the AAAI Conference on Artificial Intelligence}, publisher={Association for the Advancement of Artificial Intelligence (AAAI)}, author={Khattar, Vanshaj and Jin, Ming}, year={2023}, pages={14286–14294} }

@inproceedings{lazic2018data,
 author = {Nevena Lazic and Craig Boutilier and Tyler Lu and Eehern J. Wong and Binz Roy and M. Ryu and Gregory P. Imwalle},
 booktitle = {Neural Information Processing Systems},
 pages = {3818-3827},
 title = {{Data center cooling using model-predictive control}},
 year = {2018}
}

@book{nocedal2006numerical, title={{Numerical Optimization}}, ISBN={9780387400655}, ISSN={2197-1773}, DOI={10.1007/978-0-387-40065-5}, journal={Springer Series in Operations Research and Financial Engineering}, publisher={Springer New York}, author={Nocedal, Jorge and Wright, Stephen J.}, year={2006} }

@incollection{nweye2024chesca, title={{Winning the 2023 CityLearn Challenge: A Community-Based Hierarchical Energy Systems Coordination Algorithm}}, ISBN={9781643685489}, ISSN={1879-8314}, DOI={10.3233/faia240757}, booktitle={ECAI 2024}, publisher={IOS Press}, author={I. Garmendia, Andoni and Morri, Francesco and Cappart, Quentin and Le Cadre, Hélène}, year={2024} }

@article{oldewurtel2012use, title={{Use of model predictive control and weather forecasts for energy efficient building climate control}}, volume={45}, ISSN={0378-7788}, DOI={10.1016/j.enbuild.2011.09.022}, journal={Energy and Buildings}, publisher={Elsevier BV}, author={Oldewurtel, Frauke and Parisio, Alessandra and Jones, Colin N. and Gyalistras, Dimitrios and Gwerder, Markus and Stauch, Vanessa and Lehmann, Beat and Morari, Manfred}, year={2012}, pages={15–27} }

@book{borrelli2017predictive, title={{Predictive Control for Linear and Hybrid Systems}}, ISBN={9781107652873}, DOI={10.1017/9781139061759}, publisher={Cambridge University Press}, author={Borrelli, Francesco and Bemporad, Alberto and Morari, Manfred}, year={2017} }

@article{ji2023safety,
  publtype={informal},
  author={Jiaming Ji and Borong Zhang and Jiayi Zhou and Xuehai Pan and Weidong Huang and Ruiyang Sun and Yiran Geng and Yifan Zhong and Juntao Dai and Yaodong Yang},
  title={{Safety-Gymnasium: A Unified Safe Reinforcement Learning Benchmark}},
  year={2023},
  cdate={1672531200000},
  journal={CoRR},
  volume={abs/2310.12567}
}

@article{roijers2013survey, title={{A Survey of Multi-Objective Sequential Decision-Making}}, volume={48}, ISSN={1076-9757}, DOI={10.1613/jair.3987}, journal={Journal of Artificial Intelligence Research}, publisher={AI Access Foundation}, author={Roijers, D. M. and Vamplew, P. and Whiteson, S. and Dazeley, R.}, year={2013}, pages={67–113} }

@misc{schulman2017proximal,
  doi = {10.48550/ARXIV.1707.06347},
  author = {Schulman, John and Wolski, Filip and Dhariwal, Prafulla and Radford, Alec and Klimov, Oleg},
  title = {{Proximal Policy Optimization Algorithms}},
  publisher = {arXiv},
  year = {2017},
  copyright = {arXiv.org perpetual, non-exclusive license},
  note={arXiv:1707.06347}
}

@inproceedings{skalse2022lexicographic, series={IJCAI-2022}, title={{Lexicographic Multi-Objective Reinforcement Learning}}, DOI={10.24963/ijcai.2022/476}, booktitle={Proceedings of the Thirty-First International Joint Conference on Artificial Intelligence}, publisher={International Joint Conferences on Artificial Intelligence Organization}, author={Skalse, Joar and Hammond, Lewis and Griffin, Charlie and Abate, Alessandro}, year={2022}, pages={3430–3436}, collection={IJCAI-2022} }

@InProceedings{stooke2020responsive,
  title = 	 {{Responsive Safety in Reinforcement Learning by {PID} Lagrangian Methods}},
  author =       {Stooke, Adam and Achiam, Joshua and Abbeel, Pieter},
  booktitle = 	 {Proceedings of the 37th International Conference on Machine Learning},
  pages = 	 {9133--9143},
  year = 	 {2020},
  editor = 	 {III, Hal Daumé and Singh, Aarti},
  volume = 	 {119},
  series = 	 {Proceedings of Machine Learning Research},
  publisher =    {PMLR}
}

@inproceedings{tessler2019reward,
title={{Reward Constrained Policy Optimization}},
author={Chen Tessler and Daniel J. Mankowitz and Shie Mannor},
booktitle={International Conference on Learning Representations},
year={2019},
}

@inproceedings{vazquezcanteli2019citylearn, series={BuildSys ’19}, title={{CityLearn v1.0: An OpenAI Gym Environment for Demand Response with Deep Reinforcement Learning}}, DOI={10.1145/3360322.3360998}, booktitle={Proceedings of the 6th ACM International Conference on Systems for Energy-Efficient Buildings, Cities, and Transportation}, publisher={ACM}, author={Vázquez-Canteli, José R. and Kämpf, Jérôme and Henze, Gregor and Nagy, Zoltan}, year={2019}, pages={356–357}, collection={BuildSys ’19} }

@article{wang2020reinforcement, title={{Reinforcement learning for building controls: The opportunities and challenges}}, volume={269}, ISSN={0306-2619}, DOI={10.1016/j.apenergy.2020.115036}, journal={Applied Energy}, publisher={Elsevier BV}, author={Wang, Zhe and Hong, Tianzhen}, year={2020}, pages={115036} }

@inproceedings{wei2017deep, series={DAC ’17}, title={{Deep Reinforcement Learning for Building HVAC Control}}, DOI={10.1145/3061639.3062224}, booktitle={Proceedings of the 54th Annual Design Automation Conference 2017}, publisher={ACM}, author={Wei, Tianshu and Wang, Yanzhi and Zhu, Qi}, year={2017}, pages={1–6}, collection={DAC ’17} }

@article{zangwill1967nonlinear, title={{Non-Linear Programming Via Penalty Functions}}, volume={13}, ISSN={1526-5501}, DOI={10.1287/mnsc.13.5.344}, number={5}, journal={Management Science}, publisher={Institute for Operations Research and the Management Sciences (INFORMS)}, author={Zangwill, Willard I.}, year={1967}, pages={344–358} }

@article{zhang2019whole, title={{Whole building energy model for HVAC optimal control: A practical framework based on deep reinforcement learning}}, volume={199}, ISSN={0378-7788}, DOI={10.1016/j.enbuild.2019.07.029}, journal={Energy and Buildings}, publisher={Elsevier BV}, author={Zhang, Zhiang and Chong, Adrian and Pan, Yuqi and Zhang, Chenlu and Lam, Khee Poh}, year={2019}, pages={472–490} }

@article{polese2023coloran, title={{ColO-RAN: Developing Machine Learning-Based xApps for Open RAN Closed-Loop Control on Programmable Experimental Platforms}}, volume={22}, ISSN={2161-9875}, DOI={10.1109/tmc.2022.3188013}, number={10}, journal={IEEE Transactions on Mobile Computing}, publisher={Institute of Electrical and Electronics Engineers (IEEE)}, author={Polese, Michele and Bonati, Leonardo and D’Oro, Salvatore and Basagni, Stefano and Melodia, Tommaso}, year={2023}, pages={5787–5800} }
}

\end{document}